\documentclass[journal]{vgtc}                     

\onlineid{0}

\vgtccategory{Research}

\vgtcpapertype{please specify}

\title{\texorpdfstring{ViDDAR: Vision Language Model-Based Task-Detrimental \\ Content Detection for Augmented Reality}{ViDDAR: Vision Language Model-Based Task-Detrimental Content Detection for Augmented Reality}}

\author{%
  Yanming Xiu,
  Tim Scargill,
  Maria Gorlatova
}

\authorfooter{
  \item
  	Yanming Xiu is with Duke University.
  	E-mail: yanming.xiu@duke.edu
  \item
  	Tim Scargill is with Duke University.
  	E-mail: ts352@duke.edu

  \item 
        Maria Gorlatova is with Duke University.
        E-mail: maria.gorlatova@duke.edu
  
}

\abstract{%
  In Augmented Reality (AR), virtual content enhances user experience by providing additional information. However, improperly positioned or designed virtual content can be detrimental to task performance, as it can impair users' ability to accurately interpret real-world information. In this paper we examine two types of task-detrimental virtual content: \emph{obstruction attacks}, in which virtual content prevents users from seeing real-world objects, and \emph{information manipulation attacks}, in which virtual content interferes with users' ability to accurately interpret real-world information. We provide a mathematical framework to characterize these attacks and create a custom open-source dataset for attack evaluation.  
  To address these attacks, we introduce \textbf{ViDDAR} (\textbf{Vi}sion language model-based Task-\textbf{D}etrimental content \textbf{D}etector for \textbf{A}ugmented \textbf{R}eality), a comprehensive full-reference system that leverages Vision Language Models (VLMs) and advanced deep learning techniques to monitor and evaluate virtual content in AR environments, employing a user-edge-cloud architecture to balance performance with low latency. To the best of our knowledge, ViDDAR is the first system to employ VLMs for detecting task-detrimental content in AR settings. Our evaluation results demonstrate that ViDDAR effectively understands complex scenes and detects task-detrimental content, achieving up to 92.15\% obstruction detection accuracy with a detection latency of 533 ms, and an 82.46\% information manipulation content detection accuracy with a latency of 9.62 s. 
}

\keywords{Mixed / Augmented Reality, Vision Language Models, Object Detection, Task-Detrimental Content, Scene Understanding}

\graphicspath{{figs/}{figures/}{pictures/}{images/}{./}} 

\usepackage{tabu}                      
\usepackage{booktabs}                  
\usepackage{lipsum}                    
\usepackage{mwe}                       

\usepackage{mathptmx}                  

\renewcommand\footnotemark{}

\usepackage{multirow}

\usepackage{float}

\usepackage{tabularx}

\usepackage{colortbl}

\onlineid{1307}

\vgtccategory{Research}

\usepackage{amsmath}

\usepackage{mdframed}
\usepackage{makecell}
\usepackage{url}

\begin{document}



\firstsection{Introduction}

\maketitle

\begin{figure}[t]
\includegraphics[width=0.98\linewidth]{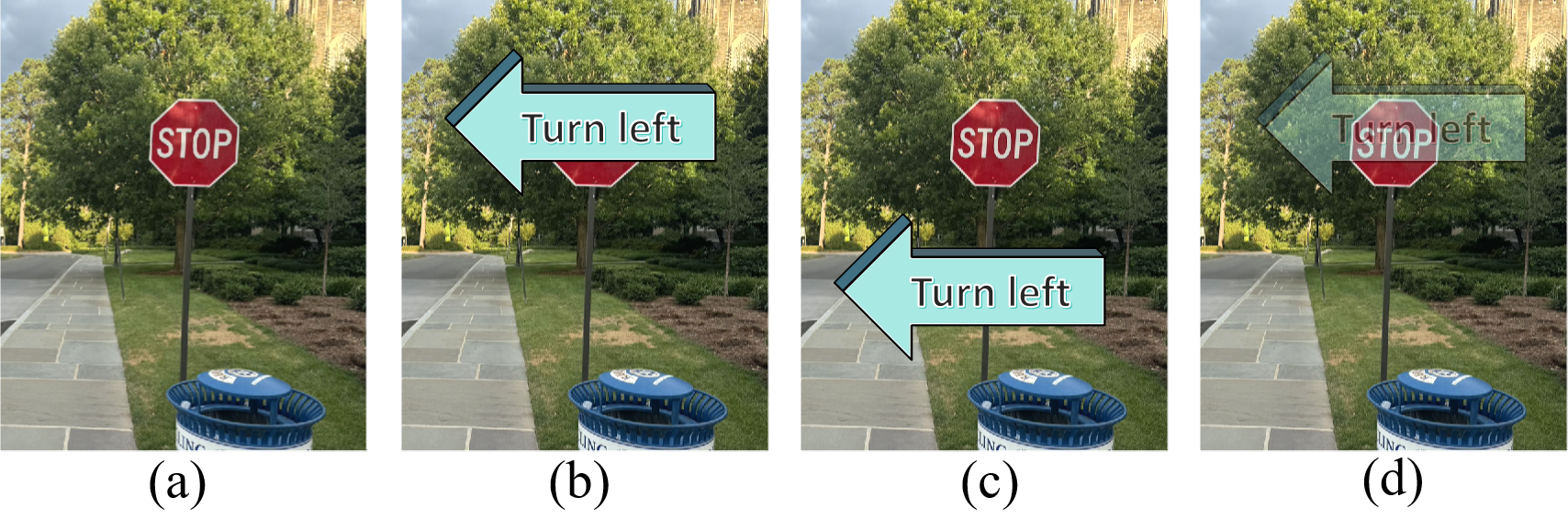}
\centering
\vspace{-0.3cm}
\caption{Example of an obstruction attack in AR: (a) real-world view; (b) AR view with a stop sign obstructed by a virtual arrow; (c) obstruction is mitigated by moving the virtual arrow; (d) obstruction is mitigated by making the virtual arrow translucent.}
\label{fig:obstruction stop}
\vspace{-0.8cm}
\end{figure}

\begin{figure}[t]
\includegraphics[width=1\linewidth]{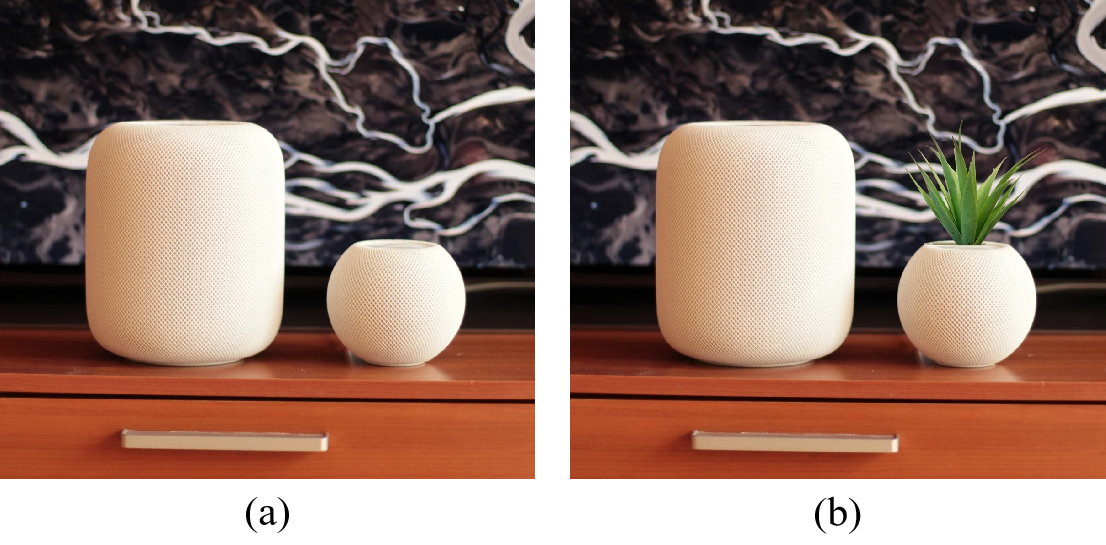}
\centering
\vspace{-0.7cm}
\caption{Example of an information manipulation attack in AR: (a) real-world view, showing two speakers; (b) AR view with a virtual plant placed on one speaker. Users may misinterpret the speaker as a flowerpot.}
\label{fig:confusion-plant-speaker}
\vspace{-0.6cm}
\end{figure}

Augmented Reality (AR) integrates virtual elements into the physical world, offering users enriched and immersive experience while providing practical assistance across various domains, including entertainment, education, and professional settings. 
However, previous studies have revealed that improperly positioned or designed content can be detrimental to task performance. These issues 
may cause users to overlook or misinterpret real-world information~\cite{attack1, attack2, misleading01}, 
leading to impaired performance on tasks that require a comprehensive understanding of the environment.
One example is the \emph{obstruction attack}, in which virtual content 
prevents users from seeing real-world objects~\cite{obstructionismar1, obstructionismar2, xiu2024lobstar, obstruction3, obstruction04}. 
This issue is particularly critical when the obstructed object is essential for task performance or user safety. 
For example, in Fig.~\ref{fig:obstruction stop}, a virtual navigation arrow obstructs a real stop sign, potentially causing the user to turn directly onto the road, leading to potential accidents. 
A more subtle and complex issue is the \emph{information manipulation attack}~\cite{misleading01, confusion}. In this scenario, virtual content is improperly designed and lowers users' ability to accurately interpret real-world information. Such attacks manipulate users' perception, leading to misunderstandings about the functionality or information of real-world elements. For instance, if a virtual plant is placed on a smart speaker, as shown in Fig.~\ref{fig:confusion-plant-speaker}, users might mistake the speaker for a flowerpot and attempt to water the plant, potentially causing damage. These attacks are challenging to evaluate because they rely not on visual overlap between virtual and real-world elements, but on the semantic interpretation of the scene, complicating detection.


To address these challenges, several methods have been proposed to assess the quality of virtual content in AR. One approach involves adapting full-reference image quality assessment (FR-IQA) methods~\cite{FRIQA, confusion}. The underlying philosophy is that better AR images should perform better in terms of FR-IQA metrics. However, most traditional algorithms~\cite{SSIM, VIF, saliency, MSSSIM} developed for FR-IQA rely on local features, such as pixel- or patch-level comparisons, limiting their capability to understand environmental information, especially in complex scenes~\cite{FRIQAlimitation}.

Furthermore, previous research has highlighted the limitations of traditional computer vision methods in simulating human perception~\cite{traditionalIQAlimitation01, traditionalIQAlimitation02, traditionalIQAlimitation03}. Human perception involves not only recognizing visual patterns but also understanding context, purpose, and relationships between objects in a meaningful way. This critical gap makes it challenging for these approaches to accurately identify critical information within an image and to interpret how virtual content interacts with or affects the real-world environment in complex scenarios. As a result, these inherent limitations can lead to unreliable assessments of virtual content quality, potentially compromising safety and diminishing the overall user experience in AR applications.

In light of these limitations, more sophisticated methods are required to handle both obstruction and information manipulation attacks. Such methods must not only detect critical visual information but also interpret the context and meaning of objects within their environment. This necessitates models that go beyond traditional algorithms to offer a more holistic and human-like analysis of the interaction between virtual and real-world content. Recent advancements in machine learning (ML), particularly in vision language models (VLMs), offer promising solutions to these challenges. Unlike traditional algorithms, VLMs integrate visual and textual information, enabling a more comprehensive and macroscopic understanding of complex scenes~\cite{vlmability01, vlmreview}. These models are highly effective at context-aware analysis, capturing intricate relationships between objects and producing detailed descriptions that closely mirror human perception. This enables VLMs not only to detect objects but also to analyze the relationships and interactions between them, making these models particularly well-suited for AR applications where understanding context is critical. As VLMs can produce human-like interpretations of scenes, they hold significant potential for improving the detection of both obstruction and information manipulation attacks in AR settings.

In this work, we present \textbf{ViDDAR} (\textbf{Vi}sion language model-based task-\textbf{D}etrimental content \textbf{D}etector for \textbf{A}ugmented \textbf{R}eality), a comprehensive full-reference system that leverages VLMs and advanced deep learning techniques to monitor and evaluate virtual content in AR environments. It is designed to analyze both raw and augmented images, enabling accurate scene
understanding and detection of task-detrimental virtual content. By providing real-time detection and actionable feedback, ViDDAR aims to enhance user experience and ensure the safe and effective use of AR applications. The key contributions of this paper are as follows:

\begin{itemize}

    \item We formally defined two categories of task-detrimental AR content: obstruction attacks and information manipulation attacks, providing a 
    mathematical framework to describe their characteristics and allowing for more precise analysis and detection.
    

    \item We proposed ViDDAR, a system that uses VLMs and other ML models to detect these attacks and assess the quality of virtual content in AR environments. ViDDAR employs a user-edge-cloud architecture to balance performance with low latency. To our knowledge, \emph{ViDDAR is the first system to employ VLMs for detecting task-detrimental content in AR settings}.


    \item We created a dataset featuring examples of both obstruction and information manipulation attacks. To validate the accuracy of the dataset labeling, we conducted a user study approved by the Duke University Campus Institutional Review Board (protocol number: 2020-0292). The results demonstrate that our labeling aligns closely with human perception. The dataset is available on GitHub.\footnote{\label{viddar-dataset}\url{https://github.com/YM-Xiu/ViDDAR-Dataset}}

    \item We evaluated ViDDAR using both datasets and real-world AR application image streams.  In detecting obstruction attacks, ViDDAR achieves up to 92.15\% accuracy with a detection latency of 533 ms on an Android mobile app. In detecting information manipulation attacks, ViDDAR achieves up to 82.46\% accuracy with a latency of 9.62 s.
    
\end{itemize}

The remaining sections in this paper are organized as follows: Section \ref{sec:related work} reviews the related work, followed by Section \ref{sec:modeling} that models two types of task-detrimental content in AR. In Section \ref{sec:system design} we describe ViDDAR's design and implementation. Section \ref{sec:evaluation} presents ViDDAR's evaluation on pre-collected datasets and a real-world AR application, as well as a user study conducted to validate the dataset labeling. We discuss the limitations and future work in Section \ref{sec:limitation} before concluding the paper in Section \ref{sec:conclusion}.
\section{Related Work}
\label{sec:related work}

\subsection{Task-Detrimental Content in AR}

\par\noindent\textbf{Obstruction Attack}: In AR, virtual content is designed to enhance the user’s interaction with the physical world. However, improperly placed or designed content can introduce challenges, leading to detrimental effects on the user’s experience. One of the earliest recognized issues was the obstruction attack, where virtual content blocks key real-world objects. In AR settings, virtual content is typically overlaid onto the real-world scene, which can inevitably result in some level of obstruction. This issue was
initially examined in studies exploring the impact of AR content on user safety and task performance~\cite{obstruction3, attack2, obstruction04}. In these studies, researchers observed that virtual content placed in the user's field of view could obstruct important real-world elements, potentially leading to dangerous situations. For example, in navigation systems, virtual content may overlap with critical signs, preventing users from seeing warnings or directions. To address these challenges, several methods have been proposed to detect and mitigate task-detrimental content in AR environments. Manisah et al.~\cite{obstruction04} proposed a model-based approach, where 3D models of real-world scenarios are pre-created to determine whether there is an obstruction. Davari et al.~\cite{obstructionismar2} proposed a system for managing obstruction in AR settings. It detects obstruction by calculating collisions between glanceable virtual content and the user’s view frustum. The system employs techniques such as translucency adjustment to maintain the visibility of real-world elements. Satkowski et al.~\cite{obstructionismar1} investigated alternative AR content placement areas, such as the ceiling and floor, to avoid obstructions in users' primary line of sight. Arya~\cite{obstruction3} detected obstruction attacks by using system sensors to identify critical real-world objects, such as humans and road signs, in a simulated AR environment. It ensures that critical items are not obstructed by modifying or removing virtual content based on predefined policies. Nonetheless, these systems rely on predefined scenes or important objects, which limits their generalizability, as they cannot dynamically adapt to new environments or contexts.

\par\noindent\textbf{Information Manipulation Attack}: While obstruction attacks in AR have received researchers' attention due to their direct impact on user safety and performance, information manipulation attacks remain a relatively underexplored area. This type of attack involves virtual content that misleads users about the nature or function of real-world objects, potentially leading to inappropriate interpretations or actions. 
Wang et al.~\cite{wang2024dark} introduced the concept of information manipulation design techniques in AR, highlighting how AR environments can deceive users through visual obfuscation and misleading interactions, influencing their perception and behavior. Eghtebas et al.~\cite{confusionattackexamples01} explored several hypothetical scenarios in which AR might deceive users by manipulating their perception of real-world elements, imagining potential consequences of AR misuse that could lead to confusion. However, these work primarily focus on conceptual exploration without providing practical detection mechanisms, leaving the issues unaddressed. CFIQA~\cite{confusion, confusion02} proposed the confusing image quality assessment model to address visual confusion in AR by assessing the perceptual quality of superimposed AR and real-world images, 
integrating both 
traditional and neural network-based methods. 
SARD~\cite{saliencyAR} used a combination of traditional saliency models~\cite{saliency} and ML-based models to evaluate the interaction between virtual content and background scenes and the information manipulation level of that interaction. However, these works create AR scenarios by artificially merging two static images, which does not fully capture the interactive and dynamic nature of typical AR experiences. Moreover, these methods focus on visual features and cannot assess information manipulation at the semantic level, which is vital for understanding how virtual content may mislead users about the functionality or information of real-world objects. In this work, we aim to address these limitations by proposing a VLM-based method to evaluate information manipulation at the semantic level, providing a deeper understanding of how virtual content interacts with real-world objects and potentially misleads users about their functionality or information. Additionally, we validate our method in real AR applications, ensuring that it captures the dynamic and interactive nature of AR experiences. 

\subsection{Vision Language Models for Cognitive Tasks}

Recent advancements in machine learning have enhanced the ability of models to simultaneously understand and interpret both visual and linguistic information, strengthening their cognitive capabilities. Radford et al.~introduced CLIP~\cite{clip}, a pre-trained model that employs transformer-based models~\cite{transformer, visualtransformer} for both its image and text encoders, facilitating a stronger connection between the two modalities by learning a unified, joint representation. Through pre-training, CLIP learns from millions of image-text pairs and enables zero-shot transfer across various computer vision tasks, including image classification and object recognition. 

While models like CLIP have demonstrated impressive capabilities in understanding and aligning visual and linguistic information, recent advancements in generative AI have taken this further by enabling models not only to understand but also to generate content across multiple modalities. Models like Claude~\cite{anthropic2024claude3}, LLaVA~\cite{liu2023llava}, Gemini\cite{geminiteam2024geminifamilyhighlycapable} and GPT-4v~\cite{GPT4V} have pushed the boundaries of multimodal learning by incorporating generative capabilities. These advancements are particularly relevant for cognitive tasks such as visual question answering, scene understanding, and multimodal reasoning, where models need to deeply comprehend and synthesize visual and linguistic information.

Inspired by the recent developments in VLMs and their cognitive abilities, in this work we leverage VLMs to evaluate and assess content quality in AR settings. By giving the models images and asking well-designed questions, we utilize the deep understanding that VLMs offer across both visual and linguistic modalities and aim to detect and analyze task-detrimental AR content such as obstruction and information manipulation attacks.

\section{Task-Detrimental Content Modeling}
\label{sec:modeling}

To systematically identify, evaluate, and mitigate these issues, formal mathematical models of task-detrimental content are needed. This section introduces two types of task-detrimental content—obstruction attacks and information manipulation attacks—and proposes models to capture their nature and characteristics.

\subsection{Obstruction Attack}
\label{subsec:obstruction}

\begin{figure}[t]
\includegraphics[width=1\linewidth]{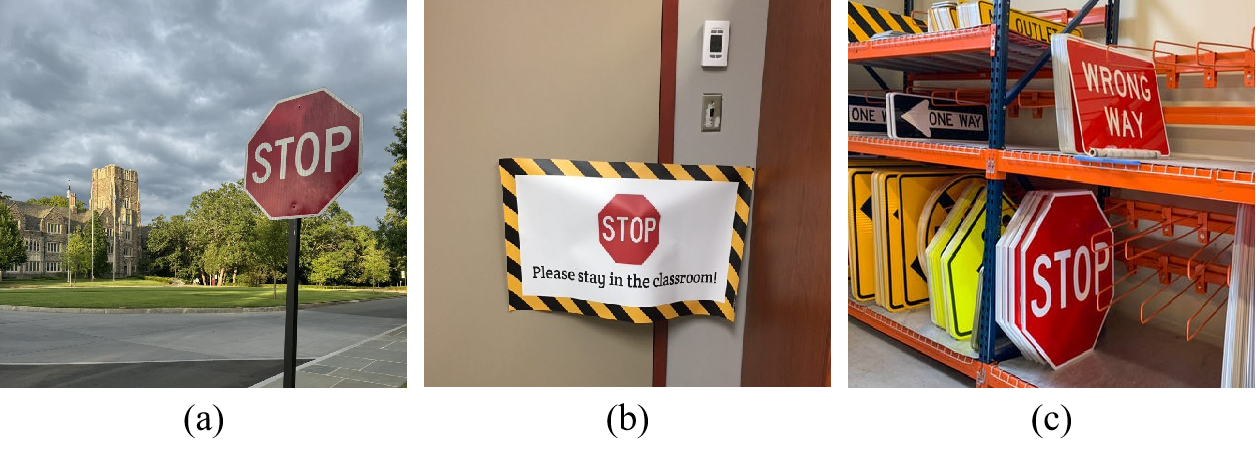}
\centering
\vspace{-0.6cm}
\caption{Whether an object is considered a key object depends on the scenario. (a): A stop sign is considered a key object when it is located on a road. (b) A stop sign is considered a key object when displayed on a door. (c): A stop sign is not considered a key object when it is a product for sale in a store.}
\label{fig:different stop sign}
\vspace{-0.6cm}
\end{figure}

We begin by modeling obstruction attacks, as they are relatively simpler and more objective to evaluate. This process aligns with the principles of full-reference image quality assessment methods\cite{full}, where comparisons are made between the raw image $I_r$ and the altered -- in our case, augmented, -- image $I_a$. While users can only see $I_a$, the system simultaneously monitors both $I_r$ and $I_a$. In $I_r$, there exists a set~$K$ containing $n$ "key objects," which are potentially important and may need users' attention. The key objects $k_i \in K$ are not predefined and can vary dynamically depending on different scenarios. For instance, while a stop sign is often considered important, it may not be treated as a key object when it appears as a product to be sold on a shelf, as Fig.~\ref{fig:different stop sign} shows.  For each key object $k_i$, there is a pixel-level mask $m_k^i$ which defines the extent of $k_i$ as a set of pixels; $|m_k^i|$ denotes the number of pixels within $m_k^i$. All pixel-level masks $m_k^i$ collectively form a set, denoted as $M_k$.
\begin{equation}
   K = \{k_1, k_2, \cdots, k_n\},  
\end{equation}
\begin{equation}
    M_k = \{m_k^1, m_k^2, \cdots, m_k^n\}. 
\end{equation}

In $I_a$, virtual content is incorporated into the scene captured in $I_r$. In many closed-source AR applications, direct access to the properties of virtual content, such as its shape, coordinates, and orientation, is not available to the user. Therefore, we represent the entire collection of virtual content by $c$. For each key object $k_i$, there is a certain relationship between $k_i$ and $c$, forming a combination ($k_i, c$). The full set of combinations is denoted as $C$. Additionally, there is a pixel-level mask $m_c$ for virtual content $c$. \begin{equation}
    C = \{(k_1, c), (k_2, c), \cdots, (k_n, c)\}. 
    \label{eq:combine}
\end{equation}

We define an obstruction attack $O$ in terms of the image pair $(I_r, I_a)$: 
\begin{equation}
    O(I_r, I_a) = \begin{cases} 
    1 & \text{if } \exists (k_i, c) \in C \longrightarrow |m_k^i \cap m_c| \geq \alpha \cdot |m_k^i| \\
    0 & \text{otherwise}
    \end{cases}, 
\end{equation} 
where $\alpha$ is a threshold value determining the obstruction level. Recall that $K$ is not predefined, and the system must dynamically determine the key objects. Consequently, 
$M_k$ is also unknown and must be inferred by the system. Meanwhile, $m_c$ can be directly extracted by comparing $I_r$ and $I_a$ at the pixel level. Thus, the problem of detecting an obstruction attack is reduced to recognizing the key objects $K$ based on the specific scenario and accurately obtaining their masks $m_k$.

\subsection{Information Manipulation Attack}
\label{subsec:information manipulation modeling}


Information manipulation attacks are more challenging to quantify, as they involve the user's interpretation of the functionality or meaning of real-world elements in the presence of virtual content. Given the subjective nature of these attacks, quantifying the level of information manipulation is challenging. Instead, we adopt a binary approach and use Boolean variables to model a number of factors that may contribute to information manipulation.

Similar to \ref{subsec:obstruction}, we let $I_r$ denote the raw image and $I_a$ denote the augmented image. The virtual content in $I_a$ is represented by $c$. There is a set of real objects in $I_a$ that is represented by $R$:  
\begin{equation}
   R = \{r_1, r_2, \cdots, r_n\},
\end{equation}
\noindent where $r_i$ represents a real object and $n$ denotes the total number of real objects in $R$. Similar to Equation \ref{eq:combine}, there are combinations between virtual content and real objects:
\begin{equation}
    C = \{(r_1, c), (r_2, c), \cdots, (r_n, c)\}.
\end{equation}

\begin{figure}[t]
\includegraphics[width=1\linewidth]{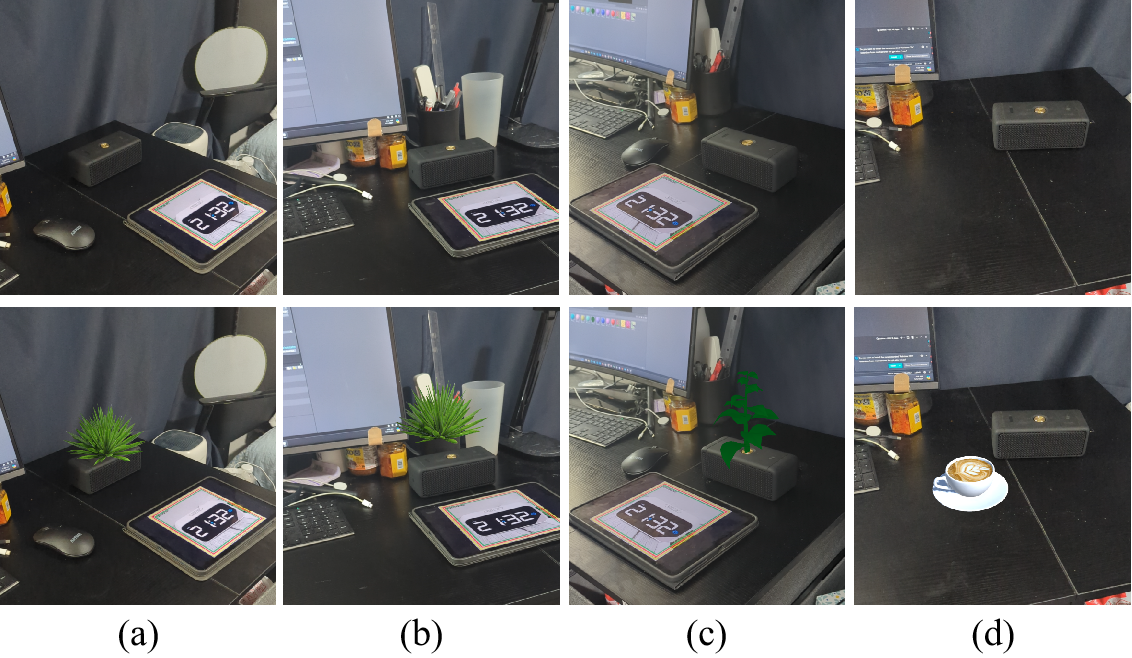}
\centering
\vspace{-0.6cm}
\caption{Images illustrating the key factors in information manipulation attacks. The top row shows raw images, while the bottom row shows augmented images. (a) A plant placed on a smart speaker can mislead the user; (b) When the plant is misaligned with the speaker, it becomes more noticeable as virtual content, reducing its potential to manipulate scene information; (c) A low-quality plant texture that does not visually blend with the real world makes the virtual content less likely to cause information manipulation; (d) Although the coffee cup is well aligned with the speaker and has a realistic style, it does not explicitly manipulate any information, as it is unlikely to misrepresent the speaker’s functionality.}
\label{fig:confusion demos}
\vspace{-0.6cm}
\end{figure}

For each of the combinations $(r_i, c)$, we use the following three Boolean variables to evaluate its level of information manipulation as perceived by users. Representative images 
that illustrate these Boolean variables are shown in Fig.~\ref{fig:confusion demos}.

\begin{itemize}
    \item \textbf{Alignment Precision}, denoted by $A$: Indicates whether the real object and the virtual content are well aligned in terms of placement or positioning. In information manipulation attacks, accurate alignment is essential for making the virtual content appear as a natural extension of the real-world object. Spatially misaligned virtual content is more likely to be recognized as virtual, making it less likely to mislead users~\cite{alignment01, alignment02}.
    
    \item \textbf{Style Similarity}, denoted by $S$: Determines whether the real object and the virtual content share a similar visual style, such as color and texture. High style similarity helps the virtual content blend seamlessly into the real-world environment, making it more difficult for users to distinguish between the two\cite{style01, style02}. Without such similarity, the virtual content would appear as an out-of-place element, making it easier for users to identify the object as virtual and reducing the potential for information manipulation.  

    \item \textbf{Information Misrepresentation}, denoted by $I$: This factor influences whether the virtual content manipulates scene information. It pertains to the extent to which the combination of virtual content and the real world causes users to misunderstand the scene. For instance, users might either overlook critical details in the scene or misinterpret non-existent information as real. 

\end{itemize}

In the context of information manipulation attacks, the variables \( A \), \( S \), and \( I \) are combined using an `AND' logic to determine whether an information manipulation attack occurs. For an attack to occur, all of $A$, $S$, and $I$ must be true—these three conditions are \textit{necessary} for the virtual content to blend seamlessly with the real object, creating the basis for information manipulation. 

Finally, based on the discussion above, we define an information manipulation attack $M$ in terms of the image pair $(I_r, I_a)$:
\begin{equation}
    M(I_r, I_a) = \begin{cases} 
    1 & \text{if } \exists (r_i, c) \in C \longrightarrow A_i \land S_i \land I_i \\
    0 & \text{otherwise}
    \end{cases},
    \label{equation:confusion}
\end{equation}
\noindent where $A_i$, $S_i$, and $I_i$ are the variables $A$, $S$ and $I$ of the combination $(r_i, c)$.

\section{ViDDAR System Design}
\label{sec:system design}

The ViDDAR architecture is deployed across three devices: an AR device, an edge server, and a cloud server. The data transmission between them is conducted using the HTTP protocol. We designed two variants of ViDDAR: one for obstruction attacks and another one for information manipulation attacks. These two variants work together to detect task-detrimental content.

\subsection{ViDDAR for Obstruction Detection}

\begin{figure}
\includegraphics[width=1\linewidth]{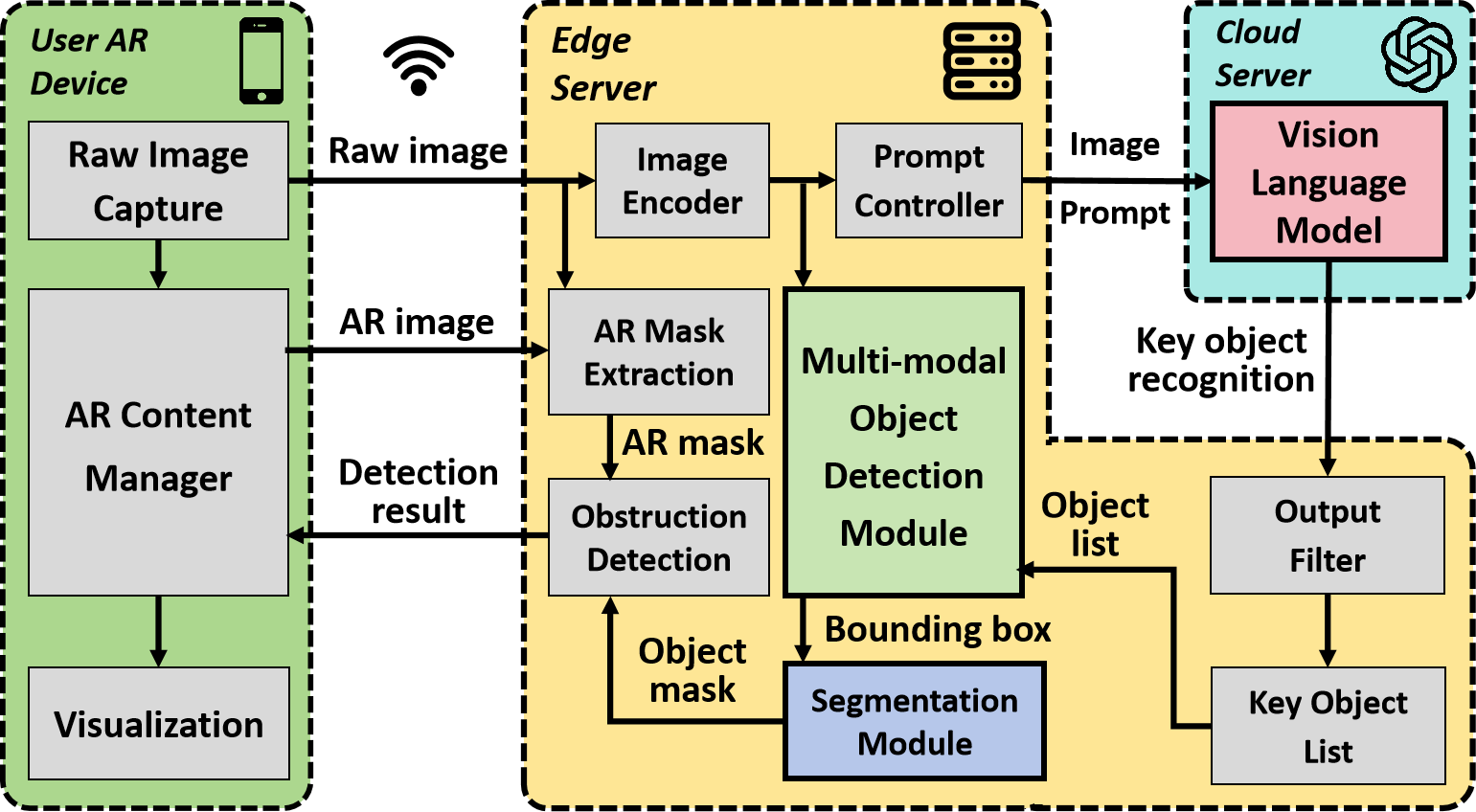}
\centering
\vspace{-0.5cm}
\caption{System architecture of ViDDAR for obstruction detection.}
\label{fig:diagram-obstruction}
\vspace{-0.65cm}
\end{figure}

The system architecture we designed for VIDDAR to detect obstruction attacks is shown in Fig.~\ref{fig:diagram-obstruction}.

\noindent \textbf{AR Device:} The AR device continuously captures raw camera images and overlays virtual content onto them. It transmits both the raw image and the augmented image to the edge server and receives obstruction detection results. If an obstruction of a key object is detected, the virtual content's opacity is reduced to ensure the key object remains visible, thereby notifying the user of the obstruction of the critical object.

\noindent \textbf{Edge Server:} The edge server receives the raw image \( I_r \) and the augmented image \( I_a \) from the AR device. The raw image is encoded in Base64 format and sent to the cloud server with a text prompt to detect the key objects in the image. This process is managed by a prompt controller to minimize resource usage, ensuring that only a small fraction of raw image frames is sent to the cloud server, 
reducing both cost and latency. At present, this process is initiated manually by the user. In the future, we plan to develop and integrate an automatic prompt scheduler to streamline this operation and further optimize efficiency. To manage the objects detected by the VLM on the cloud server, we introduce a "key object list" that stores the names of key objects, enabling ViDDAR to recognize these objects within the scene. 
Simultaneously, each raw image and its corresponding key object name list are passed to the multi-modal object detection module, which generates bounding boxes of the key objects. These bounding boxes are processed by a segmentation module to produce binary masks. Unlike the VLM prompt, object detection and segmentation are performed on every raw image. Finally, these masks are compared with virtual content masks at the pixel level, where the virtual content mask generated by comparing \( I_r \) and \( I_a \).

In our implementation, we use Grounding DINO~\cite{groundingdino} as the multi-modal object detection model. Grounding DINO is a state-of-the-art open-set object detection model. It can detect multiple objects based on a text prompt without the need to predefine the categories during training, making it ideal for ViDDAR since the category and number of key objects are both unknown. After the bounding boxes are generated, we use the Segment Anything Model (SAM)~\cite{segmentanything} to generate binary masks of key objects. As a foundation model, SAM employs zero-shot learning to generalize across various domains without requiring extensive retraining. It also supports segmenting objects inside a bounding box, allowing for seamless integration with Grounding DINO.

\noindent \textbf{Cloud Server:} 
The cloud server hosts a VLM, as deploying state-of-the-art VLMs on edge servers is challenging for most use cases due to VLMs' large number of parameters and high resource demands. Relying on cloud-based services enables us to leverage the superior performance of VLMs. The VLM processes the encoded raw image, identifies the key object within it, and sends the name of the identified object back to the edge server. In our implementation, we tested LLaVA-Next-8b\cite{liu2023llava} and GPT-4o-2024-08-06\cite{GPT4V} as the VLM.

\subsection{ViDDAR for Information Manipulation Detection}

\begin{figure}[t]
\includegraphics[width=1\linewidth]{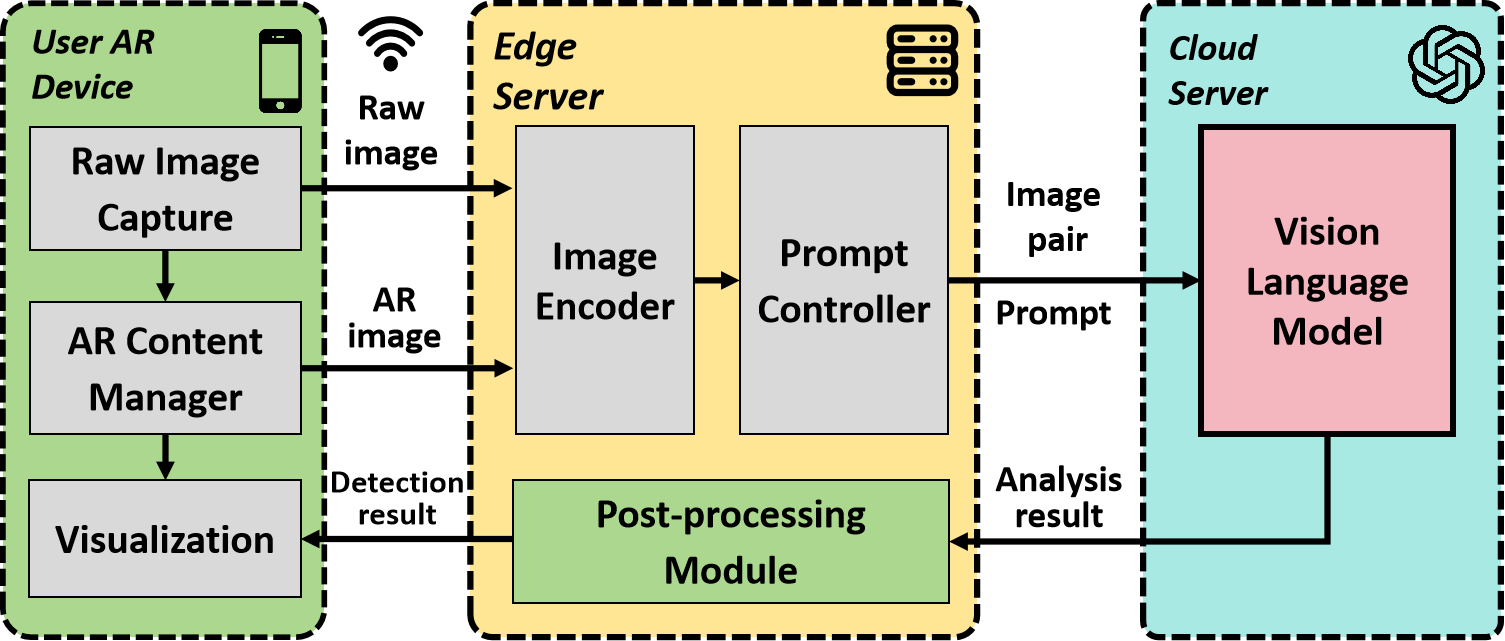}
\centering
\vspace{-0.5cm}
\caption{System architecture of ViDDAR for information manipulation detection.}
\label{fig:diagram-confusion}
\vspace{-0.65cm}
\end{figure}

The system architecture we designed for VIDDAR to detect information manipulation attacks is shown in Fig.~\ref{fig:diagram-confusion}.

\noindent \textbf{AR Device:} Similar to ViDDAR for obstruction detection, the AR device in ViDDAR for information manipulation detection also continuously captures raw images and overlays them with virtual content. It sends both \( I_r \) and \( I_a \) to the edge server for processing, receiving information manipulation detection results. When an information manipulation attack is detected, a warning message is displayed to the user.

\noindent \textbf{Edge Server:} In information manipulation attack detection, which involves more subjective cognitive evaluation, fewer modules are employed on the edge server. Instead, most of the evaluation is handled by the VLM hosted on the cloud server. The edge server receives \( I_r \) and \( I_a \) and encodes the images in Base64 format. 
To optimize resource usage, a prompt controller manages when the encoded images are sent, ensuring the VLM is only employed when necessary. In the current design, images are sent only upon user request. 
Given that the cloud server analyzes multiple factors, such as alignment and style similarity, the prompt and the VLM's output can be lengthy and complex. To simplify obtaining the detection result, a post-processing module is implemented on the edge server to interpret the VLM's output and provide a binary "True/False" decision. This is done by checking which of "yes" and "no" answers appeared closest to the end of the VLM's text response.

\noindent \textbf{Cloud Server:} The cloud server employs a VLM to detect information manipulation attacks by analyzing both $I_r$ and $I_a$. This dual-input approach enables the VLM to evaluate the interaction between real-world scenes and virtual content, effectively identifying nuanced information manipulation attacks that alter users' perception of real-world objects.

\begin{figure*}
\includegraphics[width=1\linewidth]{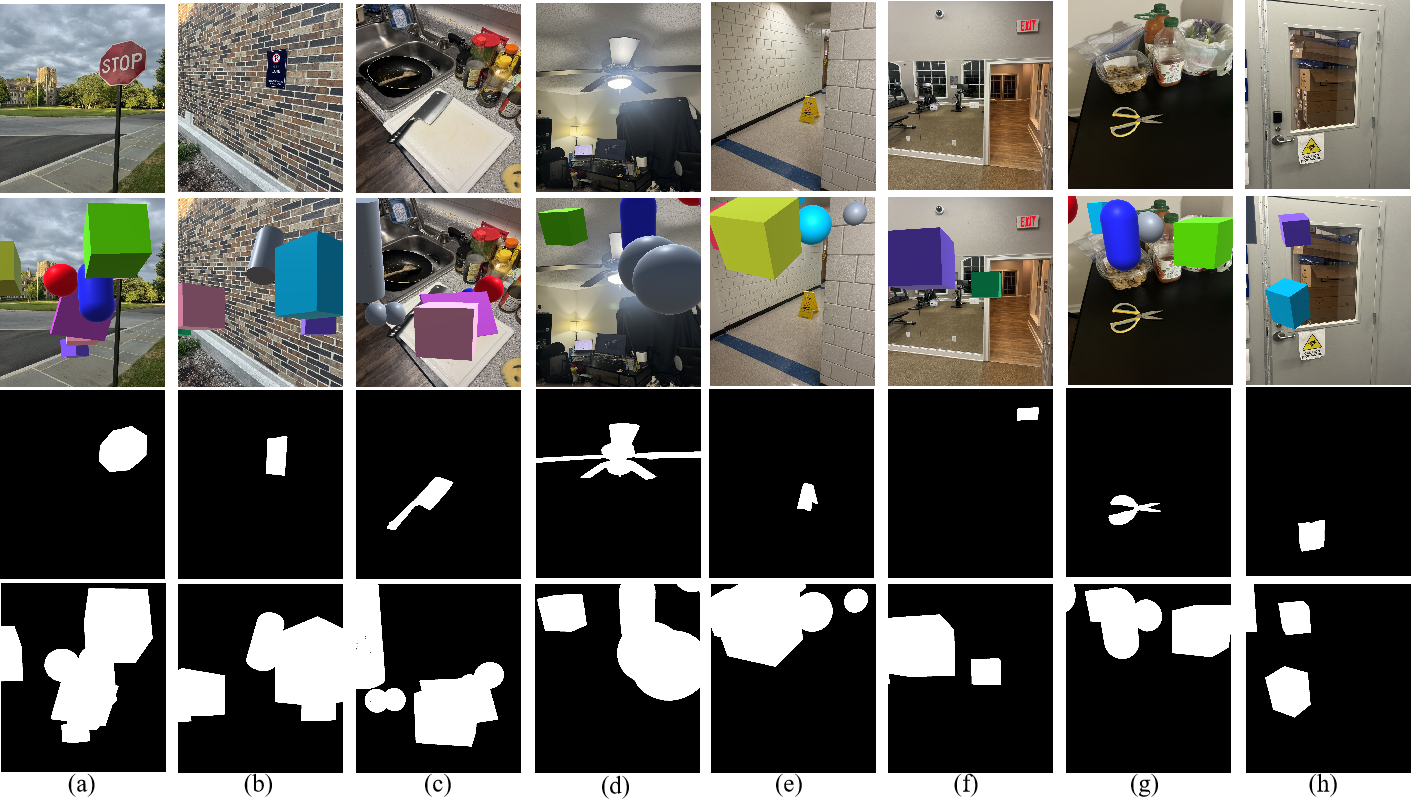}
\centering
\vspace{-0.8cm}
\caption{Obstruction attack dataset samples. The first row shows raw images; the second row shows the augmented images; the third row shows the ground truth key object mask; the fourth row shows the virtual content mask. The key objects in each column are: (a): stop sign; (b): "no parking" sign; (c): knife; (d): ceiling fan; (e): caution sign; (f): exit sign; (g): scissors; (h): biohazard sign. Data in columns (a-d) are labeled as "obstruction," while those in columns (e-h) are labeled as "no obstruction." }

\label{fig:obstruction_data_sample}
\vspace{-0.5cm}
\end{figure*}

\section{System Evaluation}

\label{sec:evaluation}


We evaluated ViDDAR using two approaches: (1) pre-collected image datasets and (2) real-time image streams. To collect images for the datasets, we developed an Android-based AR application using Unity and Google ARCore. For obstruction attack detection, we randomly placed various geometric shapes within the AR environment to create obstruction scenarios. For information manipulation attack detection, which requires precise alignment between virtual content and real-world objects, we relied on ARCore's image tracking functionality. Specifically, we introduced a physical calibration image—a marker displayed on an iPad screen, measuring 15x15 cm—into the scene. The calibration image was tracked by ARCore to establish a spatial anchor, enabling the precise positioning of virtual content relative to the calibration image. This ensured consistent alignment for evaluating information manipulation attacks. The same application was also used to evaluate ViDDAR’s performance in real-time image streams.

We used a Google Pixel 7 Pro as the AR device, and the AR app was developed with Unity 2022.3.28f1. The edge server had three NVIDIA GeForce RTX 3090. The cloud server was accessed through the OpenAI API for GPT-4o-2024-08-06 and the Hugging Face API for LLaVA-Next-8b.

\subsection{Obstruction Detection Results on the Dataset}

\subsubsection{Experiment Setup}

\begin{figure}[t]
\includegraphics[width=1\linewidth]{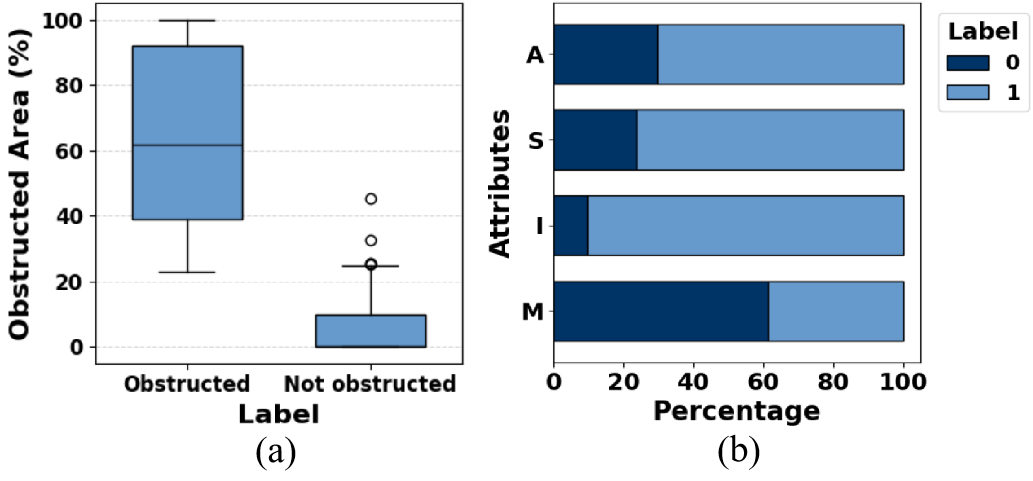}
\centering
\vspace{-0.7cm}
\caption{Statistical analysis of datasets we collected: (a): Distribution of the key object's obstructed area percentage for "obstructed" and "not obstructed" images in the obstruction dataset; (b): Label distribution percentages for attributes $A$ (alignment precision), $S$ (style similarity), $I$ (information misrepresentation), and $M$ (information manipulation) in the information manipulation dataset.}
\label{fig:statistics}
\vspace{-0.6cm}
\end{figure}

We collected a dataset for evaluating ViDDAR's performance in obstruction attack detection tasks.
The dataset contains 306 image pairs $(I_r, I_a)$ from real-world environments, with one key object in each image pair and a total of 23 classes of key objects across the entire dataset. 
We manually labeled the key object class, binary mask, and obstruction status for the image pairs. Representative image samples from the dataset are shown in Fig.~\ref{fig:obstruction_data_sample}. The percentage of obstructed area of the dataset images is shown in Fig.~\ref{fig:statistics}(a). During analysis, we identified that the approximate boundary between "obstructed" and "not obstructed" images was 0.25, so we set the obstruction threshold $\alpha=0.25$. The dataset is publicly available on GitHub\footref{viddar-dataset}.


We crafted the following prompt for key object recognition, using strategies including role assignment\cite{roleprompt}, few-shot prompting\cite{fewshotprompt}, and fine-grained formatting\cite{GPT4V}. This design ensured that the VLM focused on identifying only the most critical objects in each image, with a 
bias toward safety-related or attention-demanding elements.

\begin{mdframed}
"\textbf{You are an expert in observing the world.} \textbf{Based on the scenario,} identify the key object that needs people’s attention or safety inspection in the image based on the scenario. \textbf{Give only one object} that you think is important to be noticed, and do not provide any other information. \textbf{The objects can be caution information signs, electrical devices, safety equipment, etc.} If you think the color is important, \textbf{you can also mention the color}, such as `red box,' but be precise and describe the object with no more than 4 words."
\end{mdframed}




\begin{table*}[ht]
\caption{Obstruction attack detection results of ViDDAR and baselines.}
\vspace{-0.2cm}
\renewcommand{\arraystretch}{1.7}  
\centering
\begin{tabular}{c|ccc|c}
\hline
\textbf{Detection Method} & 
\makecell{\textbf{Vision Language}\\ \textbf{Model}} & 
\makecell{\textbf{Key Object Recognition}\\ \textbf{Accuracy (\%)}} &
\makecell{\textbf{Segmentation}\\ \textbf{Mean IoU (\%)}} &
\makecell{\textbf{Obstruction Attack }\\ \textbf{Detection Accuracy (\%)}} \\
\hline
\hline
\multirow{2}{*}{\textbf{ViDDAR}} & \textbf{GPT-4o}  & 91.83 & \underline{\textbf{72.15}} & \underline{\textbf{92.15}} \\ 
                          & LLaVA-Next-8b & 85.95 & 71.85 & 89.21 \\
\hline
\multirow{2}{*}{End-to-End} & GPT-4o  & 91.83 & N/A & 81.04 \\ 
                          & LLaVA-Next-8b & 85.95 & N/A & 55.23 \\
\hline
\multirow{2}{*}{Underdetailed} & GPT-4o  & 86.60 & 70.21 & 86.92 \\ 
                          & LLaVA-Next-8b & 76.47 & 61.58 & 81.37 \\
\hline
\multirow{2}{*}{Greedy} & GPT-4o  & \underline{\textbf{93.14}} & 67.93 & 88.89 \\ 
                          & LLaVA-Next-8b & 88.89 & 64.84 & 85.62 \\
\hline
Saliency Map & N/A & N/A & N/A & 51.63 \\
\hline
Canny Edge & N/A & N/A & N/A & 51.96 \\
\hline
Prior Knowledge & N/A & 100.00 & 78.83 & 93.14 \\
\hline

\end{tabular}
\vspace{-0.4cm}
\label{table:1}
\end{table*}

For comparison, we also implemented several baselines:


\begin{itemize}

    \item \textbf{Prior knowledge:} In this baseline, the object detection module is provided with the key object information directly, bypassing the need for VLM-based recognition. This baseline serves as a "performance upper bound" since it achieves 100\% key object recognition accuracy, thereby improving the precision of the overall object detection, segmentation, and obstruction detection processes.

    \item \textbf{End-to-end:} This baseline employs the VLM as an end-to-end solution through a two-step procedure. In the first step, the VLM is tasked with identifying the key object in \( I_r \), similar to the standard ViDDAR approach. In the second step, the VLM is provided with both \( I_r \) and \( I_a \) and is asked directly whether the key object identified in the first step is being obstructed. In essence, this approach relies on the VLM to perform the tasks typically handled by the object detection and segmentation modules. The prompt of step 2 is designed as follows, where \{key\_obj\} is the output of step 1.

    \begin{mdframed}
    "You are an expert in augmented content analysis. Look at both images. The first image is the raw image and there is a \{key\_obj\} in it. The second image is an augmented image created by overlaying some virtual content on the raw image. Identify whether the virtual elements in the second image are obstructing the \{key\_obj\}. If the \{key\_obj\} is blocked or obfuscated, then answer Yes. If the \{key\_obj\} is not blocked or obfuscated then answer No. The answer should contain only `Yes' or `No.'"
    \end{mdframed}


    \item \textbf{Underdetailed:}  This baseline follows the same pipeline as standard ViDDAR, but we prompt the VLM with minimal information and detail as follows, with no role assignment or few-shot examples.

    \begin{mdframed}
    "Identify the key object in the image. Give only one object that you think is important to be noticed. Give the name of the object only and do not provide any other information." 
    \end{mdframed}

    \item \textbf{Greedy:} To avoid missing or making mistakes in choosing key objects in the raw image, this baseline asks the VLM to find all of the potential key objects within the image and detects them individually using Grounding DINO. The prompt asks "\textit{Give any object that you think is important to be noticed}," instead of "\textit{Give only one object that you think is important to be noticed}."

    \item \textbf{Saliency map:} This baseline is based on the traditional computer vision method: saliency map~\cite{saliency}. It calculates the mean saliency score of the entire raw image $\overline{S_r}$ and the obstructed area $\overline{S_o}$. If $\overline{S_o} > \overline{S_r}$, indicating that the overlaid area is richer in information, we identify it as an obstruction, as it suggests that details are being covered by the virtual content.

    \item \textbf{Canny edge:} This baseline employs the Canny edge detection algorithm~\cite{canny} to measure the edge density in the image. It calculates the mean Canny edge score for the entire raw image $\overline{C_r}$ and the masked area where AR content is overlaid $\overline{C_o}$. If $\overline{C_o} > \overline{C_r}$, indicating that the obstructed area in the raw image has a higher edge density compared to the overall image, we classify this as an obstruction, as it suggests that details are being covered by the virtual content.

\end{itemize}

In the experiment settings above, where applicable, the multi-modal object detection model is Grounding DINO 1.5 and the segmentation module is SAM ViT-B.

\subsubsection{Results}

For our proposed method, as well as the end-to-end, underdetailed, and greedy baselines, we tested both GPT-4o and LLaVA-Next-8b as the VLM. GPT-4o is generally considered to be more powerful while LLaVA can operate with lower latency and avoid additional costs. We evaluated the performance of each method using three metrics: key object recognition accuracy, segmentation mean intersection over union (mIoU), and obstruction attack detection accuracy. Key object recognition accuracy measures the ability of the VLM to identify critical objects in the scene, while segmentation mIoU quantifies how well the system segments the key objects from the background. Obstruction attack detection accuracy is the key metric, as it directly reflects the method's ability to identify when virtual content obstructs important objects in the scene. The results are shown in Table \ref{table:1}.

Our results show that when using GPT-4o as the VLM, ViDDAR achieved the highest mIoU (72.15\%) and obstruction detection accuracy (92.15\%) if the prior knowledge baseline is excluded (as it is under ideal conditions and works as a performance upper bound). Notably, the accuracy of ViDDAR is very close to that of the prior knowledge baseline (93.14\%). When using LLaVA-Next-8b, ViDDAR also performed well, with an mIoU of 71.85\% and a detection accuracy of 89.21\%, slightly behind GPT-4o. Furthermore, ViDDAR also largely outperformed traditional computer vision-based methods such as saliency map and Canny edge in detection accuracy, which only achieved detection accuracy of 51.63\% and 51.96\%, correspondingly.

We analyze the performance of each baseline method in detail:

\begin{itemize}

    \item \textbf{Prior knowledge:} In this baseline, the object detection module is provided with the key object ground truth, ensuring 100\% key object recognition accuracy. As a result, it represents the upper bound of system performance and achieves  the highest obstruction attack detection accuracy, 93.14\%.
    
    \item \textbf{End-to-end:} This baseline relies on the VLM for both key object recognition and obstruction detection. However, determining whether the key object is obstructed requires identifying its location and estimating the proportion of the object overlaid by virtual content. 
    Although obstruction detection does not involve outputting numerical values, it remains largely a quantitative task rather than a qualitative one. VLMs often struggle with quantitative tasks and may exhibit hallucinations~\cite{hallucination01, hallucination02}, generating responses inconsistent with the input data. As a result, while this approach is straightforward, it is not ideal for obstruction attack detection. Compared to our proposed method, this baseline exhibited a decrease in obstruction detection accuracy of over 10\% with GPT-4o and over 30\% with LLaVA-Next-8b.

    \item \textbf{Underdetailed:} In the underdetailed baseline, the prompt is much simpler and less informative compared to the proposed method. It only asks the VLM to identify the key object in the image, without providing detailed instructions, example outputs, or relevant context to guide the process. This lack of specificity, clarity, and comprehensive instructions significantly hampers the VLM's ability to accurately recognize the key object. Compared to our proposed method, the underdetailed baseline exhibited a decrease in obstruction detection accuracy of over 5\% with \mbox{GPT-4o} and over 8\% with LLaVA-Next-8b.

    \item \textbf{Greedy:} This baseline instructs the VLM to output all potentially relevant key objects. This approach increases the likelihood of the key objects appearing in the outputs. As the results show, the greedy strategy achieves the highest key object recognition accuracy with both GPT-4o (93.14\%) and LLaVA-Next-8b (88.89\%). However, this strategy also introduces non-critical objects. When these objects are obstructed while the key object is not, the system still labels the scenario as "obstructed," leading to false positives. 
    Compared to our proposed method, the greedy baseline resulted in a decrease in obstruction detection accuracy of approximately 2\% with GPT-4o and 3\% with LLaVA-Next-8b.

    \item \textbf{Saliency map:} The saliency map baseline causes the system to focus on the most visually striking or attention-grabbing regions of an image, rather than the semantically important objects. The resulting accuracy, 51.63\%, is only slightly higher than random guessing (50\%), indicating that the saliency map-based method is ineffective at detecting obstruction attacks.

    \item \textbf{Canny edge:} The Canny edge detection baseline focuses on identifying edges and contours within the image, which highlights boundaries but does not capture the semantic relevance of objects. The accuracy, 51.96\% is also only slightly higher than random guessing, showing that the Canny edge-based method is not capable of detecting obstruction attacks.

\end{itemize}

\subsection{Information Manipulation Detection Results on the Dataset}

\subsubsection{Experiment Setup}

\begin{figure*}
\includegraphics[width=1\linewidth]{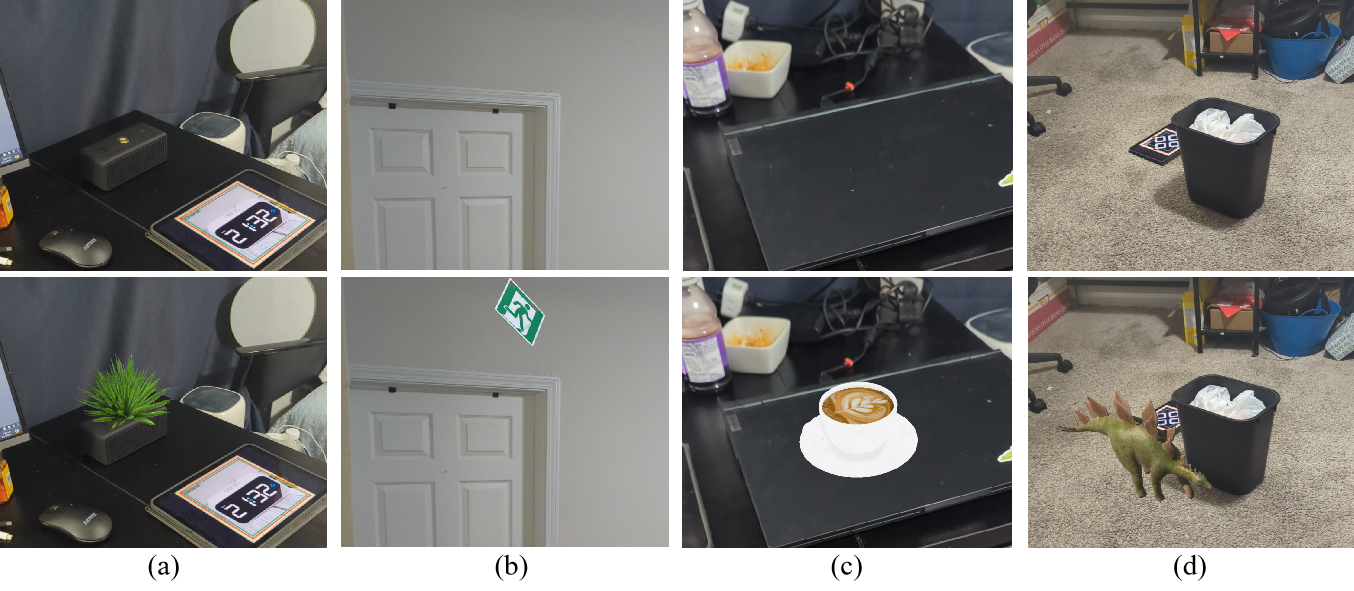}
\centering
\vspace{-0.75cm}
\caption{Information manipulation attack dataset samples. The first row shows raw images; the second row shows the augmented images. Their labels ($Alignment\ Precision$, $Style\ Similarity$, $Information\ Misrepresentation$, $Information\ Manipulation$) for each sample are as follows: (a):  A virtual plant on the speaker may lead to information manipulation, making the speaker appear to be a plant pot, with labels (1, 1, 1, 1); (b): The virtual exit sign placed above a door is potentially misleading, but it is not well aligned with the door, labeled as (0, 1, 1, 0); (c) A virtual coffee cup on a laptop may give the impression that the laptop is a food tray, potentially causing damage if other food is placed on it. However, the low-quality texture of the coffee cup makes it more noticeable, resulting in labels (1, 0, 1, 0); (d) The toy dinosaur is well aligned with the ground and the trash bin and has a high-quality texture, but it is unlikely to misrepresent the function of the trash bin, thus labeled (1, 1, 0, 0).}
\label{fig:confusion labeling}
\vspace{-0.5cm}
\end{figure*}

We created a dataset consisting of 114 image pairs $(I_r, I_a)$ from real-world scenes to evaluate ViDDAR's performance in detecting information manipulation attacks. The dataset includes 10 distinct combinations of virtual content and real-world settings, each carefully designed to reflect potential information manipulation scenarios in AR. 
For each image pair, we manually labeled the alignment precision, style similarity, information misrepresentation, and overall information manipulation status. The label distribution is shown in Fig.~\ref{fig:statistics}(b). Examples from the information manipulation dataset, along with their corresponding feature labels, are shown in Fig.~\ref{fig:confusion labeling}.

We designed the following prompt for the VLM to detect information manipulation attacks. The first two questions act as a preliminary guide, helping the VLM identify both the virtual content and the key object within the scene, establishing the necessary foundation for accurate evaluation. These initial steps ensure that the VLM is equipped to focus on relevant elements before assessing potential information manipulation.  Questions 3, 4, and 5 delve into specific attributes: $A$, $S$, and $I$ introduced in Section \ref{subsec:information manipulation modeling}.
These questions direct the VLM to evaluate whether the virtual content is properly aligned with the real-world object, visually integrated in terms of style, and whether it misrepresents or obscures the object’s original information. The final question synthesizes the responses from the previous three, providing a conclusive judgment on information manipulation attack.

\begin{mdframed}
\noindent "Here are two images. The first one is a raw image, and the second one is an augmented image, created by adding some virtual content to the space.

\noindent Please answer the following questions:

\noindent 1. What is the virtual content in the augmented image?

\noindent 2. What `key object' is interacting with the virtual content? Avoid general terms like `table surface' or `environment.'

\noindent 3. Is the virtual content accurately aligned to the object, without a significant gap? Answer yes or no, then explain why.


\noindent 4. Does the virtual content have a relatively high-quality yet reasonable texture that blends it into the real world? Answer yes or no, then explain why.

\noindent 5. Do you think the interaction will make users believe the `key object' has some false functionality or information it does not have, or lose some true functionality or information it actually has? Think creatively, only say no if the combination has no specific semantic relation. Answer yes or no, then explain why.

\noindent 6. If you answered `yes' in all questions 3, 4, and 5, you must say `yes.' Otherwise, you say `no.'"
\end{mdframed}

\subsubsection{Results}

We conducted experiments to evaluate the performance of ViDDAR on information manipulation attack detection with different choices of VLMs. Specifically, we tested GPT-4o and LLaVA-Next-8b on their ability to detect information manipulation. The results, shown in Table \ref{table:confusion}, indicate that GPT-4o achieved an accuracy of 82.46\%, demonstrating its capability to detect such attacks, while LLaVA-Next-8b performed significantly worse, with an accuracy of only 42.98\%.

In addition to accuracy, we also evaluated the models' performance in terms of precision and recall. In information manipulation attack detection, precision is of particular importance, as a false positive (incorrectly labeling content as manipulated) can unnecessarily disrupt the user’s experience. GPT-4o reached a detection precision of 74.00\% and a recall of 84.09\%, demonstrating a more balanced performance. On the other hand, LLaVA-Next-8b exhibited a high detection recall of 95.35\% but a low precision of 39.80\%, indicating that although it correctly identified information manipulation attacks when they occurred, it struggled to accurately distinguish between cases with and without information manipulation. In fact, we observed that LLaVA tends to label most of the samples as "information manipulated," highlighting its limitations in precise information manipulation content detection and resulting in lower overall performance.

\begin{table}[ht]
\caption{Information manipulation attack detection results with VLMs.}
\renewcommand{\arraystretch}{1.7}
\centering
\begin{tabular}{c|c|c|c}
\hline
\makecell{\textbf{Vision Language}\\ \textbf{Model}} & 
\makecell{\textbf{Detection}\\ \textbf{Accuracy}} &
\makecell{\textbf{Detection}\\ \textbf{Precision}} &
\makecell{\textbf{Detection}\\ \textbf{Recall}}
\\ \hline \hline
GPT-4o & \underline{\textbf{82.46\%}} & \underline{\textbf{74.00\%}} & 84.09\% \\ \hline
LLaVA-Next-8b  & 42.98\% & 39.80\% & \underline{\textbf{95.35\%}} \\ \hline
\end{tabular}
\label{table:confusion}
\vspace{-0.4cm}
\end{table}

\subsection{User Study: Dataset Labeling Validation}

\subsubsection{Study Setup}

We validated the labeling of our datasets via an IRB-approved user study, which evaluated whether the labeled key objects and information manipulation factors aligned with users' perceptions of virtual content interactions with real-world objects. We assessed users' agreement with our labeling using a custom Likert scale-based questionnaire~\cite{likert}.  
Users were provided with a questionnaire in Jupyter Notebook format, where they reviewed images on a computer monitor and recorded their responses. The study was conducted in a hybrid manner, allowing users to complete it either remotely or in person in our lab. In-person participants completed the questionnaire on a laptop we provided, 
with the displayed images measuring 
approximately 10×10 cm. 
Remote participants completed the questionnaire using a display device of their choice.

\begin{figure}[t]
    \centering
    \includegraphics[width=1\linewidth]{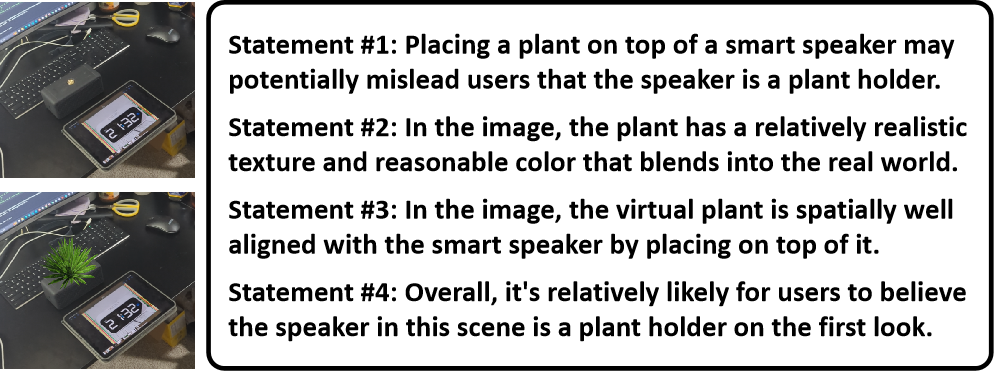}
    \vspace{-0.5cm}
    \caption{An example of statements used in the information manipulation dataset validation study. The four statements correspond to information misrepresentation ($I$), style similarity ($S$), alignment precision ($A$), and overall manipulation labeling ($M$), respectively.}
    \label{fig:user study example}
    \vspace{-0.7cm}
\end{figure}

The study involved two tasks: obstruction labeling and information manipulation labeling. For each task, participants were shown 10 image pairs and asked to provide feedback using a 5-point Likert scale, where 1 corresponds to "Strongly Disagree" and 5 corresponds to "Strongly Agree." In obstruction dataset validation, each image pair had two related statements. The participants were asked to rate their agreement with the statements, which specified the key object and obstruction. The first statement was in the format of  "The key object in the raw image is [xx]," where [xx] is the key object label of the image; the second one was in the format of "In the augmented image, [xx] is fully or partially obstructed by virtual content." In information manipulation dataset validation, each image pair was accompanied by four statements, corresponding to the three information manipulation factors and the overall manipulation status. These statements were designed based on the labels of the information manipulation factors. The first statement corresponded to information misrepresentation ($I$), which assessed the semantic potential of the combination to manipulate scene information. To avoid preconceptions, users provided their response to the first statement before reviewing the image pair. Then they responded to the remaining statements based on their evaluation of the image pair.  An example of our statement design for the information manipulation attack dataset is shown in Fig.~\ref{fig:user study example}. Among the statements, some of them were presented in a negative tone, meaning that agreeing with the statement indicated lower accuracy in labeling. To ensure consistent interpretation across all responses, we reversed the negatively-keyed scores by subtracting them from 6. This normalization step allowed us to align higher scores with higher labeling accuracy across all items.



\subsubsection{Results}

\begin{figure}[t]
    \centering
    \includegraphics[width=1\linewidth]{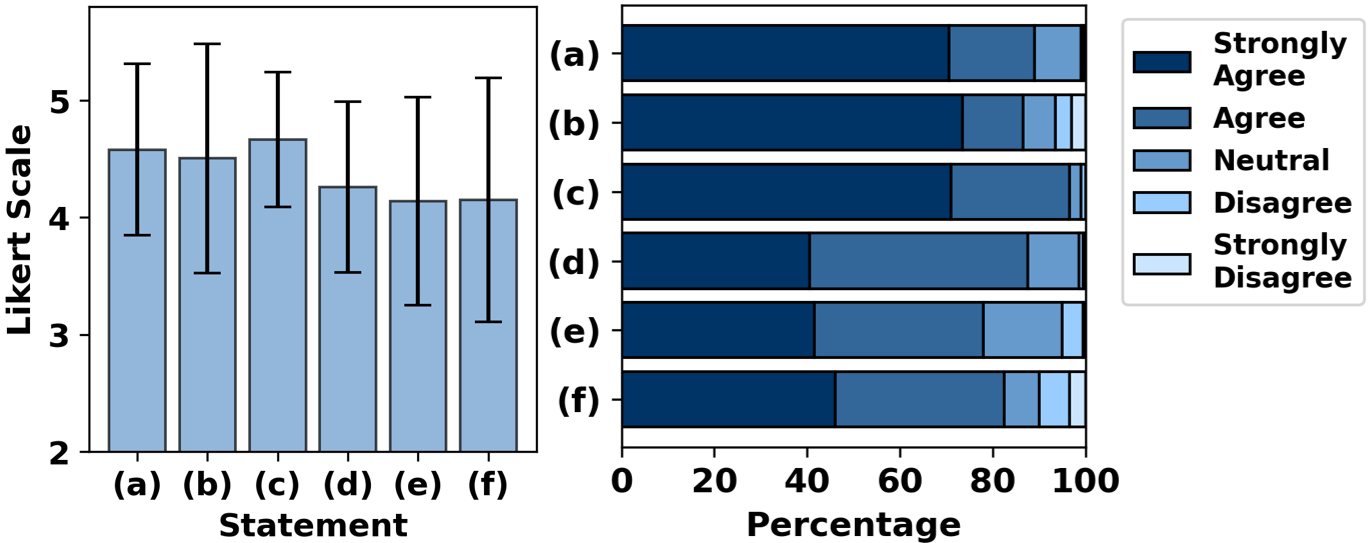}
    \vspace{-0.5cm}
    \caption{Likert scale evaluation results from the user study. Left: average agreement scores for six categories—(a) key object labeling, (b) obstruction labeling, (c) alignment precision, (d) style similarity, (e) information misrepresentation, and (f) overall manipulation labeling. Right: percentage breakdown of agreement scores, ranging from "Strongly Agree" to "Strongly Disagree," across the same categories.}
    \label{fig:likert}
    \vspace{-0.2cm}
\end{figure}

We conducted the study with 20 participants, aged 18 to 55 years, including 3 females. Among them, 16 participants completed the study on-site, while 4 participated remotely. The results are shown in Fig.~\ref{fig:likert}. From the average Likert scale histogram, all categories received an average score higher than 4 (agree). In the response percentage breakdown, we can further see that the majority of responses fell between 4 and 5, indicating a high level of agreement with our labeling. It is worth noting that key object recognition, obstruction labeling, and alignment precision labeling received higher scores, likely due to the more objective nature of these tasks. On the other hand, style similarity, information misrepresentation, and manipulation labeling showed slightly more variation, 
likely because these tasks involve more subjective judgments, which can vary between participants. Overall, the results confirm that our labeling provides a reliable dataset for evaluating systems designed to detect task-detrimental content in AR environments.

\subsection{Real-time AR Application Test}


We tested ViDDAR on our Android AR app in real-time scenarios. To evaluate latency, we recorded the smartphone screen and collected video data. In obstruction attack detection, we first prompted the VLM to initialize the key object list and measured the time between the occurrence of the obstruction and the system's response. In information manipulation attack detection, we measured the time interval from when the user initiated a detection request to when the smartphone displayed the detection result. For all tests, we used the standard ViDDAR prompt and selected GPT-4o, a commercial VLM by OpenAI, due to its superior performance. The GPT-4o model was accessed via the OpenAI API through our on-campus network in Durham, NC, USA. We evaluated two network configurations: one in which both the phone and the edge server were connected to the same on-campus network, and another in which the edge server remained on the on-campus network while the phone was connected to a student apartment network. Both configurations used a 5GHz Wi-Fi (802.11ac) connection. For each attack type and network configuration, we conducted 20 trials. The test results are shown in Table \ref{table:latency}.

The results indicate that ViDDAR can achieve a latency as low as 533 ms for obstruction detection in a one-hop network setting, allowing for rapid detection without obvious delays. In contrast, information manipulation content detection takes considerably longer, with a latency of 9.62 seconds in a one-hop network, due to the detection system's heavy reliance on VLMs. However, unlike obstruction attacks, which can 
change in real time depending on the user's spatial position and viewing pose, 
information manipulation attacks tend to remain relatively stable over time. This means that detection is not required for every frame, and such checks only need to be performed occasionally. While the increased latency may limit the applicability of this approach in scenarios that require real-time attack detection, it remains suitable for applications where periodic evaluations of information manipulation are sufficient.



\section{Limitations and Future Work}

\label{sec:limitation}

While ViDDAR has demonstrated promising results, several limitations need to be addressed. Some of them are related to the use of VLMs. Firstly, we observed 
inherent randomness in the responses of the VLM, which occasionally introduces instability in the system’s performance. Additionally, the time required to prompt the VLM on a cloud server is typically over 6 seconds. Although this delay is moderate for VLMs, it can disrupt real-time applications, especially in scenarios that require rapid responses. To address these concerns, we will continue monitoring the development of VLMs and identify models that are robust while lightweight enough for deployment on edge servers\cite{edgellm}. Fine-tuning a smaller VLM such as NVILA~\cite{liu2024nvilaefficientfrontiervisual} on AR-specific data and making it capable of attack detection can also be a solution. Finally, for information manipulation attack detection, ViDDAR currently relies heavily on the VLM to evaluate the properties of virtual content and real-world scenes, resulting in 
lower accuracy compared to that of obstruction detection. To improve this, we will refine our mathematical modeling of information manipulation attacks and introduce additional detection modules. These enhancements will help make ViDDAR more robust, ensuring better detection performance in both information manipulation and obstruction attacks.

In addition to the VLM-related improvement directions discussed above, we plan to extend ViDDAR to the analysis of more dynamic AR content. While the current work explores various types of AR content, its focus on images and static AR content may not sufficiently capture the inherently dynamic nature of AR experiences, where both virtual content and the real-world environment can continuously change. Recent advancements in video question answering~\cite{NEURIPS2023_f22a9af8} offer valuable insights that could be leveraged to enhance the analysis of dynamic scenes in future iterations of ViDDAR.


Lastly, we plan to deploy ViDDAR on additional AR platforms. Our current efforts focus on integrating ViDDAR with head-mounted devices (HMDs); as part of this research, we have already demonstrated the feasibility of employing ViDDAR with AR applications running on the Meta Quest~3. Our near-term goal is to fully implement the system and conduct comprehensive studies to evaluate its performance and impact on user experience. 

\begin{table}[t]
\caption{Detection latency of ViDDAR under different network settings.}
\vspace{-0.2cm}
\renewcommand{\arraystretch}{1.6}  
\centering
\begin{tabular}{c|c|c}
\hline
\textbf{Detection Task} & 
\makecell{\textbf{Network Settings}} & 
\makecell{\textbf{Mean Latency}} \\
\hline
\hline
\multirow{2}{*}{\makecell{Obstruction\\ Detection}} & \makecell{One-hop}  & 533 ms \\ \cline{2-3}
                          & \makecell{Six-hop} & 960 ms \\
\hline
\multirow{2}{*}{\makecell{Information \\ Manipulation\\ Detection}} & \makecell{One-hop}  & 9.62 s \\ \cline{2-3}
                          & \makecell{Six-hop} & 12.30 s \\
\hline
\end{tabular}
\label{table:latency}
\vspace{-0.5cm}
\end{table}

\section{Conclusion}

\label{sec:conclusion}

 In this work, we introduced ViDDAR, the first system to leverage VLMs for detecting task-detrimental virtual content in AR scenes. By employing both edge and cloud servers, ViDDAR achieves a balance between detection accuracy and latency. ViDDAR was rigorously tested on a pre-collected dataset, demonstrating promising detection accuracy. Additionally, we evaluated ViDDAR using real-time image streams through an Android-based AR application. To enable accurate detection, we mathematically modeled two types of AR attacks,  obstruction attacks and information manipulation attacks, providing a formal framework to assess their impact on user experience. Furthermore, the effectiveness of our dataset was validated through an IRB-approved user study. This work lays a foundation for applying VLMs to AR content evaluation and enhances user experience by promoting the safe and effective use of AR applications.

\acknowledgments{%
    We thank Ajay Divakaran and Yunye Gong for their suggestions on this paper, and Junfeng Lin for assisting with dataset image collection. We also thank our user study participants for their invaluable assistance in this research. This work was supported in part by NSF grants CSR-2312760, CNS-2112562, and IIS-2231975, NSF CAREER Award IIS-2046072, NSF NAIAD Award 2332744, a CISCO Research Award, a Meta Research Award, Defense Advanced Research Projects Agency Young Faculty Award HR0011-24-1-0001, and the Army Research Laboratory under Cooperative Agreement Number W911NF-23-2-0224. The views and conclusions contained in this document are those of the authors and should not be interpreted as representing the official policies, either expressed or implied, of the Defense Advanced Research Projects Agency, the Army Research Laboratory, or the U.S. Government. This paper has been approved for public release; distribution is unlimited. No official endorsement should be inferred. The U.S.~Government is authorized to reproduce and distribute reprints for Government purposes notwithstanding any copyright notation herein.
}










\end{document}